%% file: main.tex
\documentclass[11pt,fleqn]{article}

\usepackage{palatino,geometry}

\input{IntroDLinputsbasic.tex}

\usepackage[english]{babel}
\usepackage{caption}
\usepackage{subcaption}

\newtheorem{theorem}{Theorem}

\title{Smoothed Q-learning\\\vspace{5mm} \small{University College London Centre for AI Research Note}}
\author{David Barber}

\renewcommand{\beq}{\begin{equation}}
\renewcommand{\eeq}{\end{equation}}
\newcommand{\tQ}{\tilde{Q}}
\newcommand{\tQs}{Q_*}
\newcommand{\sQ}{Q^\pi}

\newcommand{\figref}[1]{figure(\ref{#1})}

\newcommand{\secref}[1]{section(\ref{#1})}
\renewcommand{\eqref}[1]{equation(\ref{#1})}
\newcommand{\thref}[1]{theorem(\ref{#1})}

\usepackage{mdframed}
\usepackage{caption}

\begin{document}

\maketitle

\begin{abstract}
    In Reinforcement Learning the Q-learning algorithm provably converges to the optimal solution. However, Q-learning can also overestimate the values and thereby spend too long exploring unhelpful states. Double Q-learning is a provably convergent alternative that mitigates some of the overestimation issues, though sometimes at the expense of slower convergence. We introduce an alternative algorithm that replaces the max operation with an average, resulting also in a provably convergent off-policy algorithm which can mitigate overestimation yet retain similar convergence as standard Q-learning.   
\end{abstract}

\section{Q-learning and Overestimation}

We consider Reinforcement Learning  \cite{Sutton1998}  based on a Markov Decision Process, with states $s$, actions $a$, state transition distribution $p(s'|s,a)$ and (stochastic) reward $r(s'|s,a)$ and discounting $\gamma<1$. The optimal value $Q_*(s,a)$ of taking action $a$ from state $s$  is the solution to the equation
\beq
Q_*(s,a) = \sum_{s'}p(s'|s,a)\br{r(s'|s,a) + \gamma\max_a Q_*(s',a)}
\label{eq:opt}
\eeq
Q-learning is an off-policy approach to finding $\tQs$ based on the update
\beq
Q_{t+1}(s_t,a_t) = Q_t(s_t,a_t) + \alpha_t(s_t,a_t)\br{r_t + \gamma \max_a Q_t(s_{t+1},a)-Q_t(s_t,a_t)}
\label{eq:ql:update}
\eeq
Here off-policy means that, provided the learning rate satisfies certain conditions and every state-action pair is explored \cite{Sutton1998} then, no matter how the states and actions are visited, the  update \eqref{eq:ql:update} converges to the optimal policy $Q_*$.\\

In \cite{dql} Hasselt points out that the maximisation operation in the update  \eqref{eq:ql:update}  means that, for stochastic rewards, the $Q_t$ values can be over-estimated, resulting in the learning process spending too long in states which are not profitable. Ideally we would replace these $Q_t$ values by their average over many episodes (since \eqref{eq:opt} is an optimality condition for the expected $Q_*$ over infinitely many episodes). Double Q-learning \cite{dql} addresses this overestimation by estimating two values $Q_A(s,a)$, $Q_B(s,a)$ and deciding on actions for system A based on the values for B and vice-versa. \\

Whilst double Q-learning can reduce the time spend in unprofitable states, it can also converge more slowly (though there are analyses for the linear case that suggest that this can be overcome by doubling the learning rate and taking the final $(Q_A(t)+Q_B(t))/2$ as the estimate of the Q-value \cite{NEURIPS2020_4bfbd52f}). Other approaches similarly attempt to reduce variance in the estimation of the $Q_t$ for example by calculating a running average, see  \cite{selfcorrectingQ}. However, as far as we are aware, all previous works do not replace the max operation, which can also be considered to be part of the reason for the over-estimation in Q-learning. Our modest contribution is an approach that replaces the max operation with a smoother average operation. 

%However, the Double Q-learning approach can actually also underestimate the Q values; it also requires storing two tables $Q_A(s,a)$, $Q_B(s,a)$.

\section{Smoothed Q-learning}
We define Smoothed Q-learning as
\beq
\tQ_{t+1}(s_t,a_t) = \tQ_t(s_t,a_t) + \alpha_t(s_t,a_t)\br{r_t + \gamma \sum_a q_t(a|s_{t+1})\tQ_t(s_{t+1},a)-\tQ_t(s_t,a_t)}
\label{eq:vql:update}
\eeq
where $q_t(a|s_{t+1})$ is a distribution. When $q_t(a|s_{t+1})$ places all its mass in the state $a$ that maximises $\tQ_t(s_{t+1},a)$, then this update is equivalent to standard Q-learning; otherwise it `smooths' the delta distribution and places mass in more than the most likely state. Based on the simple fact that the average is less than the maximum, we have
\beq
\sum_a q_t(a|s_{t+1})\tQ_t(s_{t+1},a) \leq \max_a \tQ_t(s_{t+1},a) 
\eeq
%where we have equality if $q_t(a|s_{t+1})$ places all its mass on the action that maximises $\tQ_t$. 
This means that the single step update of smoothed Q-learning will give a lower update value to the maximal state than standard Q-learning, potentially therefore avoiding overconfident estimates of the Q-value.\\

The update \eqref{eq:vql:update} is reminiscent of the SARSA algorithm \cite{Rummery:1994,Sutton1998} -- an on-policy algorithm to estimate the value of a policy (the actions taken by the system follow a defined policy $\pi(a|s)$)
\beq
\sQ_{t+1}(s_t,a_t) = \sQ_t(s_t,a_t) + \alpha_t(s_t,a_t)\br{r_t + \gamma \sQ_t(s_{t+1},a)-\sQ_t(s_t,a_t)}
\label{eq:svql:update}
\eeq
Whilst SARSA is an on-policy approach that converges to the value of the policy $\pi$, smoothed Q-learning is an off-policy approach that converges to the optimal value (provided we choose a $q_t$ that increasingly concentrates on the most likely action -- see below).

\subsection{Smoothing Choices}

We want to define $q_t(a|s')$ that, as $t$ increases, $q_t$ places increasing mass on the maximum value state $a^*=\arg\max_a \tQ_t(s',a)$. That is, we define distributions $q_t(a|s')$ that place mass $1-\delta_t$ in the maximal state $a^*$  with $\delta_t$ decreasing to zero as $t$ increases. Provided we choose a smoothing distribution that increasingly concentrates its mass on the maximal state, then smoothed Q-learning converges to the optimal value (see \secref{sec:convergence} for a proof sketch). There are many smoothing choices and we describe here some natural ones: 

\begin{description}
\item[Softmax]
The distribution
\beq
q_t(a|s_{t+1}) \propto e^{\beta_t \tQ_t(s_{t+1},a)}
\eeq
sharpens around the maximal state as $\beta_t$ increases with time.

\item[Clipped Max]
Another simple choice is to place $1-\delta_t$ mass in the most likely state and spread the remaining $\delta_t$ mass uniformly in the non-maximal states, decreasing $\delta_t$ towards zero as $t$ increases (assuming there are $A$ actions in state $s_{t+1}$):
\beq
q_t(a^*|s_{t+1})  = 1-\delta_t, \ocm q_t(b|s_{t+1}) = \delta_t/(A-1)
\eeq
where $a^*=\argmax{a}{\tQ(s_{t+1},a)}$, and $b$ is any state other than $a^*$.
\end{description}
Other choices which take into account the number of times a state-action pair has been visited would also be worthy of future investigation.

\begin{figure}[t]
\centering
\includegraphics[width=6cm]{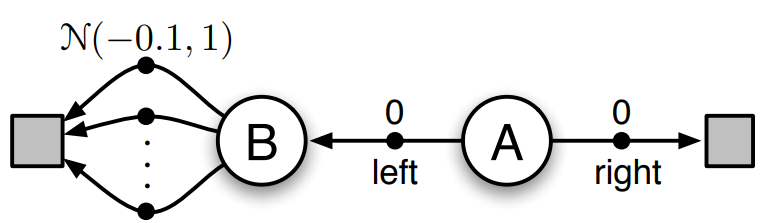}
\caption{Experiment set up in example 6.7 of \cite{Sutton1998}. The starting state is \texttt{A}. For action \texttt{Right}, the next state is the terminal state \texttt{C}, with reward 0. For action \texttt{Left} the next state is state \texttt{B} with reward 0. From state \texttt{B} there are 8 actions, each taking to terminal state \texttt{D}, with reward sampled from a Gaussian distribution with mean $-0.1$ and variance 1.\label{fig:max:bias}}
\end{figure}

\section{Demonstration}

\begin{figure}[h]
\centering
\begin{subfigure}[b]{0.485\textwidth}
\centering
\includegraphics[width=\textwidth]{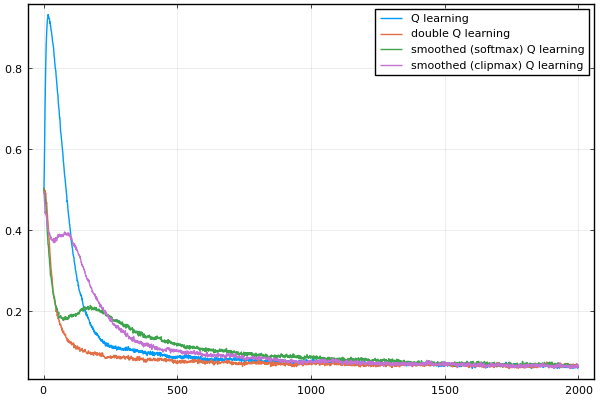}
\caption{}
\end{subfigure}
\hfill
\begin{subfigure}[b]{0.485\textwidth}
\centering
\includegraphics[width=\textwidth]{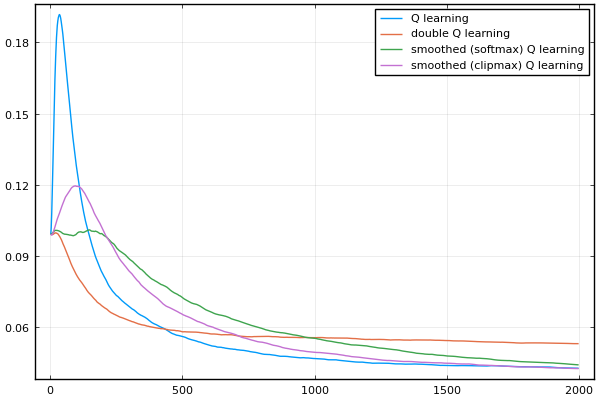}
\caption{}
\end{subfigure}
\caption{Maximisation Bias example from \cite{Sutton1998}. We plot the fraction of times the action taken from state \texttt{A} is \texttt{Left} (which is suboptimal).  (b) We plot the convergence of $|\tilde{Q} - Q_*|$ averaged over all states and actions.  As we can see, smoothed Q learning suffers less from over estimating the Q value than standard Q-learning.  Double Q-learning also resolves some of the issues with overestimation of the Q-value, but converges more slowly to the optimal value.}
\label{fig:experiment}
\end{figure}

The experiment comes from the maximisation bias example in \cite{Sutton1998},  see \figref{fig:max:bias}.  The starting state is \texttt{A}, with actions \texttt{Left} (goes to state \texttt{B}) and \texttt{Right} (terminates). Both actions have zero reward.  From \texttt{B} there are 8 actions, each of which takes the model to the terminal state, each action having reward sampled from a Gaussian distribution with mean -0.1 and variance 1. This example is challenging for standard Q-learning since, because of the high variance in the reward, going \texttt{Left} from \texttt{A} will likely result in some positive rewards (whereas the optimal action is always to go \texttt{Right} from \texttt{A}). For this reason, Q-learning can spend a long time taking Left actions from \texttt{A} until it overcomes the stochastic nature of the reward. In \figref{fig:experiment} we plot the results of using Q-learning, double Q-learning, and smoothed Q-learning (with the softmax and clipped max distributions). For all methods, we used $\epsilon=0.1$ greedy exploration, $\gamma=0.99$, and learning rate $\alpha_t = 0.1/(1+0.001 t)$. The results plotted are the averages over 10,000 experiments.  For the softmax we used $\beta_t = 0.1 + 0.1(t-1)$, and for the clipped max, we used $\delta_t = \exp(-0.02t)$.\\
 
The example shows that double Q-learning avoids some of the issues with Q-learning and pays less attention to initial over-estimates of the Q value (due to unhelpful random rewards that confuse the model to thinking that going left is a good strategy).

\section{Conclusion}

The overestimation issue of standard Q-learning is potentially addressable by replacing the max operation in the standard Q-learning by an average. We showed that this converges to the optimal Q-value and that the method has some empirical support based on a standard toy problem from the literature.

\subsubsection*{Acknowledgement}

I'd like to thank Francisco Melo for helpful conversations.

\appendix

\section{Convergence\label{sec:convergence}}

We here sketch a proof that makes use of the following result (see \cite{singh2000} for a more formal statement of the result) and follows the line of thought in \cite{melo}:
\begin{theorem}
The random process $\cb{\Delta_t}$ taking value in $R$ and defined as
\beq
\Delta_{t+1}(x) = (1-\alpha_t(x))\Delta_t(x) + \alpha_t(x)F_t(x)
\label{eq:delta:update}
\eeq
converges to 0 with probability 1 under the following assumptions:
\begin{itemize}
    \item $0\leq \alpha_t \leq 1, \hcm \sum_t \alpha_t(x)=\infty, \hcm \sum_t \alpha_t^2(x)<\infty$;
    \item $\wmn{\ave{F_t(x)}} \leq \kappa \wmn{\Delta_t} + c_t$,  \hspace{1cm} $\kappa\in[0,1)$  and $c_t$ convergences to 0 with probability 1;
    \item $var(F_t(x))\leq C(1+\wmn{\Delta_t})^2$, \hspace{1cm} $C>0$ 
\end{itemize}
where $\wmn{\Delta}$ denotes a weighted max norm. 
\label{theorem:two}
\end{theorem}
We are interested convergence of $\tQ_t$ towards the optimal value $\tQs$ and therefore define 
\beq
\Delta_t = \tQ_t(s_t,a_t) - \tQs(s_t,a_t)
\eeq
It is convenient to write the smoothed update as
\beq
\tQ_{t+1}(s_t,a_t) = \tQ_t(s_t,a_t) + \alpha_t(s_t,a_t)\br{r_t + \gamma \av{\tQ_t(s_{t+1},a)}_a-\tQ_t(s_t,a_t)}
\label{eq:vql:update:simple}
\eeq
where $\av{f(x)}_x$ means expectation of the function $f(x)$ with respect to the distribution of $x$.  Using the smoothed update, we can write
\begin{align}
\Delta_{t+1} &= \tQ_{t+1}(s_t,a_t) - \tQs(s_t,a_t)\\
& = \tQ_{t}(s_t,a_t) +\alpha_t\br{r_t + \gamma\av{\tQ(s_{t+1},a)}_{a} - \tQ_t(s_t,a_t)}- \tQs(s_t,a_t)\\
%& = \Delta_t +\alpha_t\br{r_t + \gamma\av{\tQ(s_{t+1},a)}_{a} - \tQ_t(s_t,a_t) + \tQs(s_t,a_t)-\tQs(s_t,a_t)}\\
%& = \Delta_t +\alpha_t\br{r_t + \gamma\av{\tQ(s_{t+1},a)}_{a} - \Delta_t - \tQs(s_t,a_t)}\\
& = (1-\alpha_t)\Delta_t +\alpha_t\br{r_t + \gamma\av{\tQ(s_{t+1},a)}_{a}  - \tQs(s_t,a_t)}
\end{align}
In terms of \thref{theorem:two}, we therefore define
\beq
F_t\equiv r_t + \gamma\sum_a q_t(a|s_{t+1})\tQ_t(s_{t+1},a) - \tQs(s_t,a_t)
\eeq
%For the condition in \thref{theorem:one} on the mean of $F_t$ to hold, using \eqref{eq:qv:inv:opt} to replace $\tQs(s_t,a_t)$ in the above equation, we need
For convergence, we need to bound the norm of the expected value of $F_t$. Using the fixed-point of the optimal value $\tQs$ we can write
\beq
\frac{1}{\gamma}\ave{F_t} = \avp{p(s_{t+1}|s_t,a_t)}{\sum_a q_t(a|s_{t+1})\tQ_t(s_{t+1},a) -\max_a \tQs(s_{t+1},a)}
\eeq
Defining
\beq
D_t = \sum_a q_t(a|s_{t+1})\tQ_t(s_{t+1},a) -\max_a \tQs(s_{t+1},a)
\eeq
we can form the bound
\beq
\frac{1}{\gamma}\maxnorm{\ave{F_t}}  =\maxnorm{\ave{D_t}}\leq \ave{\maxnorm{D_t}} =\maxnorm{D_t} 
\eeq
which means that if we can bound $\maxnorm{D_t}$ appropriately, the mean criterion will be satisfied.\\

%Since $\sum_a q_\infty(a|s_{t+1})\tQs(s_{t+1},a) = \max_a \tQs(s_{t+1},a)$, and 
Assuming that $q_t$ places $(1-\delta_t)$ mass in the maximal state of $\tQ$ we can write
%\beq
%\sum_a q_t(a|s_{t+1})\tQ_t(s_{t+1},a) -\sum_a q_\infty(a|s_{t+1})\tQs(s_{t+1},a)
%\eeq
%as
\begin{align}
D_t & = \max_a \tQ_t(s_{t+1},a) -\max_a \tQs(s_{t+1},a) - \delta_t\max_a\tQ_t(s_{t+1},a) +\sum_{b\neq a} q_t(b|s_{t+1})\tQ_t(s_{t+1},b) 
\end{align}
and
\begin{align}
\maxnorm{D_t} &\leq \maxnorm{\max_a \tQ_t(s_{t+1},a) -\max_a \tQs(s_{t+1},a)} + \maxnorm{ \delta_t\max_a\tQ_t(s_{t+1},a) -\sum_{b\neq a} q_t(b|s_{t+1})\tQ_t(s_{t+1},b)}\nn\\
&\leq  \max_a\maxnorm{\tQ_t(s_{t+1},a) - \tQs(s_{t+1},a)} + \maxnorm{\delta_t\max_a\tQ_t(s_{t+1},a) -\sum_{b\neq a} q_t(b|s_{t+1})\tQ_t(s_{t+1},b)}\nn\\
&=  \maxnorm{\tQ_t(s_{t+1},a) - \tQs(s_{t+1},a)} + \maxnorm{\delta_t\max_a\tQ_t(s_{t+1},a) -\sum_{b\neq a} q_t(b|s_{t+1})\tQ_t(s_{t+1},b)}\nn\\
&=  \maxnorm{\Delta_t} + \maxnorm{\delta_t\max_a\tQ_t(s_{t+1},a) -\sum_{b\neq a} q_t(b|s_{t+1})\tQ_t(s_{t+1},b)}\nn\\
&\leq  \maxnorm{\Delta_t} +\maxnorm{\delta_t\max_a\tQ_t(s_{t+1},a) - \delta_t\tQ_t(s_{t+1},b{_-})}\nn\\
%&\leq   ||\Delta_t|| + \delta_t||\max_a\tQ_t(s_{t+1},a)|| + \delta'_t||\tQ_t(s_{t+1},b_{-}) ||\nn\\
&\leq  \maxnorm{\Delta_t} + \delta_t\br{\maxnorm{\max_a\tQ_t(s_{t+1},a)} + \maxnorm{\tQ_t(s_{t+1},b_{-})}}
\end{align}
where $b_{-}=\arg\min_{b\neq a}\tQ_t(s_{t+1},b)$ and the penultimate  line follows from the fact that only a maximum of $\delta_t$ mass can be placed in the minimal state $b_{-}$ (since $(1-\delta_t)$ mass is placed in state $a^*$). Putting this together we have
\beq
\maxnorm{\ave{F_t}}  \leq \gamma\maxnorm{\Delta_t} + \gamma\delta_t\br{\maxnorm{\max_a\tQ_t(s_{t+1},a)} + \maxnorm{\tQ_t(s_{t+1},b)}}
\eeq
where
\beq
c_t \equiv {\gamma}\delta_t\br{\maxnorm{\max_a\tQ_t(s_{t+1},a)} + \maxnorm{\tQ_t(s_{t+1},b)}}
\eeq
Since the $\tQ_t$ are bounded (which follows directly from iterating the $\tQ$ update), then the term $c_t$ converges to zero with probability 1, provided $\delta_t$ converges to 0 with probability 1. The mean criterion is therefore satisfied.\\

For the variance criterion, since the rewards are bounded,  $\tQ_t$ and  $\Delta_t$ are also bounded. This means that the variance is bounded and that the criterion must therefore be satisfied for sufficiently large $C$. A more detailed argument is to first write
\beq
F_t - \ave{F_t} = r_t -\bar{r}(s_t,a_t) + \gamma\sum_a q_t(a|s_{t+1})\tQ_t(s_{t+1},a) - \gamma\sum_{s_{t+1}}p(s_{t+1}|s_t,a_t)\sum_a q_t(a|s_{t+1})\tQ_t(s_{t+1},a)
\label{eq:del:F}
\eeq
where $\bar{r}(s_t,a_t) = \sum_{s'}p(s'|s_t,a_t)r(s'|s_t,a_t)$. \\

We can write \eqref{eq:del:F} as
\begin{align}
\Delta F_t &=r_t - \av{r_t}_{s_{t+1}} + \gamma\br{\av{\tQ_t(s_{t+1},a)}_{a}  -\av{\tQ_t(s_{t+1},a)}_{s_{t+1},a}}\\
&= \Delta r_t + \gamma\br{\av{\tQ_t(s_{t+1},a)}_{a}  - \av{\tQs(s_{t+1},a)}_a + \av{\tQs(s_{t+1},a)}_a -\av{\tQ_t(s_{t+1},a)}_{s_{t+1},a}}\\
&= \Delta r_t + \gamma\av{\tQ_t(s_{t+1},a)- \tQs(s_{t+1},a)}_{a}   -\gamma\av{\av{\tQ_t(s_{t+1},a)}_{s_{t+1}}- \tQs(s_{t+1},a)}_a
%&= \Delta r_t + \av{\tQ_t(s_{t+1},a)- \tQs(s_{t+1},a)}_{a}   -\av{\av{\tQ_t(s_{t+1},a)}_{s_{t+1}}- \tQs(s_{t+1},a)}_a
\end{align}
We can bound the variance using
\beq
var(F) = \av{\Delta F_t^2} =  \maxnorm{\av{\Delta F_t^2}}\leq \av{\maxnorm{\Delta F_t^2}}=\maxnorm{\Delta F_t}^2
\eeq
and use the triangle inequality,
\beq
\maxnorm{\Delta F_t} \leq \maxnorm{\Delta r_t} + \gamma\maxnorm{\av{\tQ_t(s_{t+1},a)- \tQs(s_{t+1},a)}_{a}} +\gamma
\maxnorm{\av{\av{\tQ_t(s_{t+1},a)}_{s_{t+1}}- \tQs(s_{t+1},a)}_a}
\eeq
and using $\maxnorm{\av{x}}\leq \maxnorm{x}$
\begin{align}
\maxnorm{\Delta F_t} &\leq \maxnorm{\Delta r_t} + \gamma\maxnorm{\Delta_t} +
\gamma\maxnorm{\av{\tQ_t(s_{t+1},a)}_{s_{t+1}}- \tQs(s_{t+1},a)}%\\
%&\leq \maxnorm{\Delta r_t} + \gamma\maxnorm{\Delta_t} +
%\gamma\maxnorm{\av{\tQ_t(s_{t+1},a)}_{s_{t+1}}- \tQs(s_{t+1},a)}
\end{align}
We now write
\begin{align}
&\maxnorm{\av{\tQ_t(s_{t+1},a)}_{s_{t+1}}- \tQs(s_{t+1},a)}\\
&=\maxnorm{\av{\tQ_t(s_{t+1},a)}_{s_{t+1}}- \av{Q_*(s_{t+1},a)}_{s_{t+1}} + \av{\tQs(s_{t+1},a)}_{s_{t+1}} - \tQs(s_{t+1},a)}\\
&\leq \maxnorm{\Delta_t} + \maxnorm{\av{\tQs(s_{t+1},a)}_{s_{t+1}} - \tQs(s_{t+1},a)}\\
&\leq \maxnorm{\Delta_t} + B
\end{align}
for some constant $B_1$ since the optimal $\tQs$ is bounded (for $\gamma<1$ and bounded rewards). Hence, since the rewards are bounded, there exists $B$ such that
\beq
\maxnorm{\Delta F_t} \leq 2\gamma B +2\gamma\maxnorm{\Delta_t}=2\gamma B(1+\wmn{\Delta_t})
\eeq
for suitably defined weighted max norm. This shows that the variance condition is satisfied.

\bibliographystyle{unsrt}
\bibliography{refs}

\end{document}

%% file: IntroDLinputsbasic.tex
\usepackage{amssymb,amsmath}
\usepackage{tikz}
\usetikzlibrary{arrows,shapes,backgrounds,through,shadows,decorations}
\usetikzlibrary{decorations.pathmorphing}
\usetikzlibrary{calc}
\usetikzlibrary{matrix}

\graphicspath{{./images/}{./matlab/}{./figures/}{../matlab/}{../matlab/figures/}{../cup_first_edition/figures/}{./../codeOO/misc/}{../cup_first_edition_corrections/figures/}{./../codeOO/DemosExercises/}{../../projects/deepbelief/}
{../../talks/CambSciFest/figures/}{../../talks/CambSciFest/images/}}

\tikzstyle{bigdot}=[circle,draw=blue!80,fill=blue!25,anchor=center,align=center,minimum size=6mm]
\tikzstyle{bigemptydot}=[circle,draw=blue!80,anchor=center,align=center,minimum size=6mm]

\tikzstyle{biggreendot}=[circle,draw=green!80,fill=green!25,anchor=center,align=center,minimum size=6mm]
\tikzstyle{bigreddot}=[circle,draw=red!80,fill=red!25,anchor=center,align=center,minimum size=6mm]

\tikzstyle{dot}=[circle,draw=orange!80,fill=orange!25,anchor=center,align=center,minimum size=4mm]
\tikzstyle{emptydot}=[circle,draw=orange!80,anchor=center,align=center,minimum size=4mm]

\tikzstyle{reddot}=[circle,draw=red!80,fill=red!25,anchor=center,align=center,minimum size=4mm]

\tikzstyle{greendot}=[circle,draw=green!80,fill=green!25,anchor=center,align=center,minimum size=4mm]

\tikzstyle{bluedot}=[circle,draw=blue!80,fill=blue!25,anchor=center,align=center,minimum size=4mm]

\tikzstyle{neur}=[rectangle,draw=green!50,fill=green!50,minimum size=6mm,line width=2pt,>=stealth]  % continuous

\tikzstyle{fact}=[fill,minimum size=1.5mm,line width=2pt,>=stealth]

\tikzstyle{cont2}=[circle,draw=black!50,top color=white, % a shading that is white at the top...
bottom color=lilla!50!black!20,thick,minimum size=6mm,line width=1pt,>=stealth]  % continuous  node

\tikzstyle{contredb}=[circle,draw=red,top color=red, % a shading that is white at the top...
bottom color=red!50!black!20,thick,minimum size=8.1mm,line width=1pt,>=stealth]  % continuous  node
\tikzstyle{contyellowb}=[circle,draw=yellow,top color=yellow, % a shading that is white at the top...
bottom color=yellow!50!black!20,thick,minimum size=8.1mm,line width=1pt,>=stealth]  % continuous  node
\tikzstyle{contblueb}=[circle,draw=blue,top color=blue, % a shading that is white at the top...
bottom color=blue!50!black!20, thick,minimum size=8.1mm,line width=1pt,>=stealth]  % continuous  node

\tikzstyle{contred}=[circle,draw=red,top color=red, % a shading that is white at the top...
bottom color=white,thick,minimum size=6mm,line width=1pt,>=stealth]  % continuous  node
\tikzstyle{contyellow}=[circle,draw=yellow,top color=yellow, % a shading that is white at the top...
bottom color=yellow!50!black!20,thick,minimum size=6mm,line width=1pt,>=stealth]  % continuous  node
\tikzstyle{contblue}=[circle,draw=blue,top color=blue!80!black!50, % a shading that is white at the top...
bottom color=white, thick,minimum size=6mm,line width=1pt,>=stealth]  % continuous  node
\tikzstyle{contgreen}=[circle,draw=green,top color=green!80!black!50, % a shading that is white at the top...
bottom color=white, thick,minimum size=6mm,line width=1pt,>=stealth]  % continuous  node
\tikzstyle{contwhite}=[circle,draw=white,color=white, thick,minimum size=6mm,line width=1pt,>=stealth]  % continuous  node
\tikzstyle{contwhiteb}=[circle,draw=white,color=white, thick,minimum size=7.5mm,line width=1pt,>=stealth]  % continuous  node

\tikzstyle{cont}=[circle, draw,% a shading that is white at the top...
thick,minimum size=6mm,line width=1pt,>=stealth]  % continuous  node

\tikzstyle{contb}=[circle,draw=black!50,top color=white, % a shading that is white at the top...
bottom color=darkgreen!50!black!20,thick,inner sep=0pt,minimum size=8.1mm,line width=1pt,>=stealth]  % continuous  node

\tikzstyle{ocont}=[ellipse,draw=blue!50,thick,minimum size=6mm,>=stealth]  % continuous  node
\tikzstyle{blackcont}=[circle,draw=black!50,thick,minimum size=6mm,line width=2pt,>=stealth]  % continuous  node
\tikzstyle{oval}=[ellipse,draw=blue!50,thick,minimum size=6mm,line width=1pt,>=stealth]  % continuous node
\tikzstyle{ovalb}=[ellipse,draw=blue!50,thick,minimum size=7.5mm,line width=1pt,>=stealth]  % continuous node
\tikzstyle{disc}=[rectangle,draw=blue!50,thick,line width=1pt,minimum size=6mm]  % discrete node
\tikzstyle{obs}=[fill=blue!20,thick]  % observed node
\tikzstyle{opt}=[star,draw=red!50,thick,minimum size=6mm]  % decision node

\tikzstyle{fillred}=[fill=red!20,thick]  % observed node
\tikzstyle{fillgreen}=[fill=green!20,thick]  % observed node
\tikzstyle{purered}=[fill=red]  % observed node
\tikzstyle{state}=[rectangle,fill=red!20]  % state
\tikzstyle{sobs}=[fill=green!15,thick]  % sequentally observed node
\tikzstyle{fact}=[fill,minimum size=1.5mm,line width=2pt,>=stealth]
\tikzstyle{varfact}=[draw,minimum size=1.5mm,line width=2pt,>=stealth]
\tikzstyle{sep}=[rectangle,draw=magenta!50,thick,minimum size=6mm]  % discrete node
\tikzstyle{det}=[fill=red!15,rectangle,draw=red!50,thick,minimum size=6mm]  % deterministic node

\tikzstyle{dethid}=[diamond,draw=red!50,thick,minimum size=6mm]  % deterministic  hidden node

\tikzstyle{lineball}=[fill,-*,draw=red!50,line width=1.5pt]
\tikzstyle{redball}=[mark=*,mark options={fill=red!50,draw=red},mark size=0.5pt]
\tikzstyle{greenball}=[mark=*,mark options={fill=green!50,draw=green},mark size=0.5pt]
\tikzstyle{hid}=[circle,draw,thick]  %  non observed node

\tikzstyle{dec}=[rectangle,draw=red!50,thick,minimum size=6mm]  % decision node
\tikzstyle{utility}=[diamond,draw=red!50,thick,minimum size=6mm]  % utility node
\tikzstyle{contdec}=[circle,draw=blue!50,thick,fill=blue!10,line width=2pt]  % observed node after a decision
\tikzstyle{decutility}=[diamond,draw=red!50,thick,minimum size=6mm]  % utility node

\tikzstyle{contobs}+=[cont]
\tikzstyle{contobs}+=[obs]
\tikzstyle{discobs}+=[disc]
\tikzstyle{discobs}+=[obs]

\tikzstyle{obsred}+=[obs]
\tikzstyle{obsred}+=[red]

\tikzstyle{background grid}=[draw, black!50,step=.1cm]
\tikzstyle{dgraph}=[->, line width=1.5pt]
\tikzstyle{ugraph}=[line width=1.5pt]

\definecolor{magenta}{cmyk}{0.1,1,1,0.5}
\definecolor{darkgreen}{cmyk}{0.6,0.1,0.6,0.6}
\definecolor{pink}{cmyk}{0.1,1,1,0.1}
\definecolor{azzurro}{cmyk}{0.9333, 0.2471, 0.5569, 0.102}
\definecolor{lilla}{rgb}{0.5, 0, 0.5}
\definecolor{mygreen}{rgb}{0, 0.5, 0}
\definecolor{darkorange}{rgb}{1, 0.4, 0} % is probably not visible...
\definecolor{darkred}{rgb}{0.8, 0, 0}

\newcommand{\eframe}{\end{frame}}
\newcommand{\bframe}[1]{\begin{frame}{#1}}

\newcommand{\bmp}[1]{\begin{minipage}{#1}}
\newcommand{\bmpp}[2]{\begin{minipage}[#1]{#2}}
\newcommand{\emp}{\end{minipage}}
\newcommand{\cb}[1]{\left\{ {#1} \right\}}
\newcommand{\br}[1]{\left( {#1} \right)}
\newcommand{\sq}[1]{\left[ {#1} \right]}

\newcommand{\wmn}[1]{\left|\!\left| {#1} \right|\!\right|_W}
\newcommand{\maxnorm}[1]{\left|\!\left| {#1} \right|\!\right|_\infty}

\newcommand{\avp}[2]{\mathbb{E}_{#1}\sq{#2}}
\newcommand{\ave}[1]{\mathbb{E}\sq{#1}}
\newcommand{\av}[1]{\left\langle{#1}\right\rangle}
\newcommand{\argmax}[1]{\ensuremath{\underset{#1}{\mathrm{argmax}\ }}}

\newcommand{\ocm}{\hspace{1cm}}
\newcommand{\hcm}{\hspace{0.25cm}}

\newcommand{\beq}{\[}
\newcommand{\eeq}{\]}
\newcommand{\nn}{{\nonumber}}

\tikzstyle{celim}=[circle,draw=red!25,thick,minimum size=6mm,line width=2pt,>=stealth]  %
\tikzstyle{delim}=[draw=red!25]  %

\newcommand{\btz}{\begin{tikzpicture}}
\newcommand{\etz}{\end{tikzpicture}}

%% file: main.bbl
\begin{thebibliography}{1}

\bibitem{Sutton1998}
R.~S. Sutton and A.~G. Barto.
\newblock {\em Reinforcement Learning: An Introduction}.
\newblock The MIT Press, second edition, 2018.

\bibitem{dql}
H.~Hasselt.
\newblock {Double Q-learning}.
\newblock In J.~Lafferty, C.~Williams, J.~Shawe-Taylor, R.~Zemel, and
  A.~Culotta, editors, {\em Advances in Neural Information Processing Systems},
  volume~23. Curran Associates, Inc., 2010.

\bibitem{NEURIPS2020_4bfbd52f}
W.~Weng, H.~Gupta, N.~He, L.~Ying, and R.~Srikant.
\newblock {The Mean-Squared Error of Double Q-Learning}.
\newblock In H.~Larochelle, M.~Ranzato, R.~Hadsell, M.F. Balcan, and H.~Lin,
  editors, {\em Advances in Neural Information Processing Systems}, volume~33,
  pages 6815--6826. Curran Associates, Inc., 2020.

\bibitem{selfcorrectingQ}
R.~Zhu and M.~Rigotti.
\newblock {Self-correcting Q-Learning}.
\newblock In {\em {Proceedings of the AAAI Conference on Artificial
  Intelligence}}, volume 35(12), pages 11185--11192, 2021.

\bibitem{Rummery:1994}
G.~Rummery and M.~Niranjan.
\newblock On-line q-learning using connectionist systems.
\newblock Technical Report CUED/F-INFENG/TR 166, Cambridge University, 1994.

\bibitem{singh2000}
S.~P. Singh, T.~Jaakkola, M.L. Littman, and C.~Szepesv{\'a}ri.
\newblock {Convergence Results for Single-Step On-Policy Reinforcement-Learning
  Algorithms}.
\newblock {\em Machine Learning}, 38(3):287--308, 2000.

\bibitem{melo}
F.~S. Melo.
\newblock {Convergence of Q-learning : A simple proof}.
\newblock {\em {Institute Of Systems and Robotics, Technical Report}}, 2001.

\end{thebibliography}
